# STAIR Captions: Constructing a Large-Scale Japanese Image Caption Dataset

Yuya Yoshikawa    Yutaro Shigeto    Akikazu Takeuchi

Software Technology and Artificial Intelligence Research Laboratory (STAIR Lab)
Chiba Institute of Technology
2-17-1, Tsudanuma, Narashino, Chiba, Japan.
{yoshikawa,shigeto,takeuchi}@stair.center

## Abstract

In recent years, automatic generation of image descriptions (captions), that is, *image captioning*, has attracted a great deal of attention. In this paper, we particularly consider generating Japanese captions for images. Since most available caption datasets have been constructed for English language, there are few datasets for Japanese. To tackle this problem, we construct a large-scale Japanese image caption dataset based on images from MS-COCO, which is called *STAIR Captions*. STAIR Captions consists of 820,310 Japanese captions for 164,062 images. In the experiment, we show that a neural network trained using STAIR Captions can generate more natural and better Japanese captions, compared to those generated using English-Japanese machine translation after generating English captions.

## 1 Introduction

Integrated processing of natural language and images has attracted attention in recent years. The Workshop on Vision and Language held in 2011 has since become an annual event[1]. In this research area, methods to automatically generate image descriptions (captions), that is, *image captioning*, have attracted a great deal of attention (Karpathy and Fei-Fei, 2015; Donahue et al., 2015; Vinyals et al., 2015; Mao et al., 2015) .

Image captioning is to automatically generate a caption for a given image. By improving the quality of image captioning, image search using natural sentences and image recognition support for visually impaired people by outputting captions as sounds can be made available. Recognizing various images and generating appropriate captions for the images necessitates the compilation of a large number of image and caption pairs.

In this study, we consider generating image captions in Japanese. Since most available caption datasets have been constructed for English language, there are few datasets for Japanese. A straightforward solution is to translate English captions into Japanese ones by using machine translation such as Google Translate. However, the translated captions may be literal and unnatural because image information cannot be reflected in the translation. Therefore, in this study, we construct a Japanese image caption dataset, and for given images, we aim to generate more natural Japanese captions than translating the generated English captions into the Japanese ones.

The contributions of this paper are as follows:

- We constructed a large-scale Japanese image caption dataset, *STAIR Captions*, which consists of Japanese captions for all the images in MS-COCO (Lin et al., 2014) (Section 3).

- We confirmed that quantitatively and qualitatively better Japanese captions than the ones translated from English captions can be generated by applying a neural network-based image caption generation model learned on STAIR Captions (Section 5).

STAIR Captions is available for download from `http://captions.stair.center`.

## 2 Related Work

Some English image caption datasets have been proposed (Krishna et al., 2016; Kuznetsova et al., 2013; Ordonez et al., 2011; Vedantam et al., 2015; Isola et al., 2014). Representative examples are PASCAL (Rashtchian et al., 2010), Flickr3k (Rashtchian et al., 2010; Hodosh et al., 2013), Flickr30k (Young et al., 2014) —an extension of Flickr3k—, and MS-COCO (Microsoft Common Objects in Context) (Lin et al., 2014).

---
[1] In recent years it has been held as a joint workshop such as EMNLP and ACL; `https://vision.cs.hacettepe.edu.tr/vl2017/`

As detailed in Section 3, we annotate Japanese captions for the images in MS-COCO. Note that when annotating the Japanese captions, we did not refer to the original English captions in MS-COCO.

MS-COCO is a dataset constructed for research on image classification, object recognition, and English caption generation. Since its release, MS-COCO has been used as a benchmark dataset for image classification and caption generation. In addition, many studies have extended MS-COCO by annotating additional information about the images in MS-COCO[2].

Recently, a few caption datasets in languages other than English have been constructed (Miyazaki and Shimizu, 2016; Grubinger et al., 2006; Elliott et al., 2016). In particular, the study of Miyazaki and Shimizu (2016) is closest to the present study. As in our study, they constructed a Japanese caption dataset called YJ Captions. The main difference between STAIR Captions and YJ Captions is that STAIR Captions provides Japanese captions for a greater number of images. In Section 3, we highlight this difference by comparing the statistics of STAIR Captions and YJ Captions.

## 3 STAIR Captions

### 3.1 Annotation Procedure

This section explains how we constructed STAIR Captions. We annotated all images (164,062 images) in the 2014 edition of MS-COCO. For each image, we provided five Japanese captions. Therefore, the total number of captions was 820,310. Following the rules for publishing datasets created based on MS-COCO, the Japanese captions we created for the test images are excluded from the public part of STAIR Captions.

To annotate captions efficiently, we first developed a web system for caption annotation. Figure 1 shows the example of the annotation screen in the web system. Each annotator looks at the displayed image and writes the corresponding Japanese description in the text box under the image. By pressing the send (送信) button, a single task is completed and the next task is started.

[2]http://mscoco.org/external/

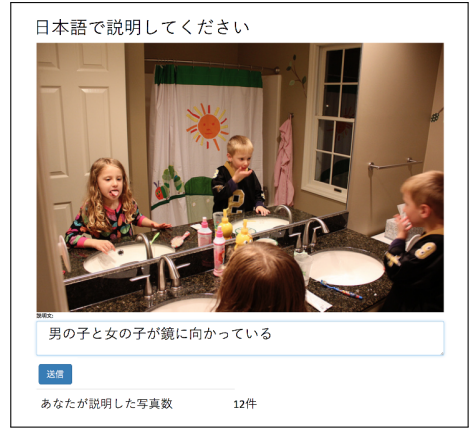

Figure 1: Example of annotation screen of web system for caption annotation.

To concurrently and inexpensively annotate captions by using the above web system, we asked part-time job workers and crowd-sourcing workers to perform the caption annotation. The workers annotated the images based on the following guidelines. (1) A caption must contain more than 15 letters. (2) A caption must follow the da/dearu style (one of writing styles in Japanese). (3) A caption must describe only what is happening in an image and the things displayed therein. (4) A caption must be a single sentence. (5) A caption must not include emotions or opinions about the image. To guarantee the quality of the captions created in this manner, we conducted sampling inspection of the annotated captions, and the captions not in line with the guidelines were removed. The entire annotation work was completed by about 2,100 workers in about half a year.

### 3.2 Statistics

This section introduces the quantitative characteristics of STAIR Captions. In addition, we compare it to YJ Captions (Miyazaki and Shimizu, 2016), a dataset with Japanese captions for the images in MS-COCO like in STAIR Captions.

Table 1 summarizes the statistics of the datasets. Compared with YJ Captions, overall, the numbers of Japanese captions and images in STAIR Captions are 6.23x and 6.19x, respectively. In the public part of STAIR Captions, the numbers of images and Japanese captions are 4.65x and 4.67x greater than those in YJ Captions, respectively. That the numbers of images and captions are large in STAIR Captions is an important point in image caption generation because it reduces the possibility of unknown scenes and objects appearing in the test images. The vocabulary of STAIR Captions is 2.69x larger than that of YJ Captions. Because of the large vocabulary of STAIR Captions, it is expected that the caption generation model can learn and generate a wide range of captions. The average numbers of characters per a sentence in STAIR Captions and in YJ Captions are almost the same.

Table 1: Comparison of dataset specifications. Numbers in the brackets indicate statistics of public part of STAIR Captions.

|  | Ours | YJ Captions |
|---:|---:|---:|
| # of images | 164,062 (123,287) | 26,500 |
| # of captions | 820,310 (616,435) | 131,740 |
| Vocabulary size | 35,642 (31,938) | 13,274 |
| Avg. # of chars | 23.79 (23.80) | 23.23 |

## 4 Image Caption Generation

In this section, we briefly review the caption generation method proposed by Karpathy and Fei-Fei (2015), which is used in our experiments (Section 5).

This method consists of a convolutional neural network (CNN) and long short-term memory (LSTM)[3]. Specifically, CNN first extracts features from a given image, and then, LSTM generates a caption from the extracted features.

Let $I$ be an image, and the corresponding caption be $Y = (y_1, y_2, \cdots, y_n)$. Then, caption generation is defined as follows:

$$\begin{aligned}
\mathbf{x}^{(im)} &= \text{CNN}(I), \\
\mathbf{h}_0 &= \tanh\left(\mathbf{W}^{(im)}\mathbf{x}^{(im)} + \mathbf{b}^{(im)}\right), \\
\mathbf{c}_0 &= \mathbf{0}, \\
\mathbf{h}_t, \mathbf{c}_t &= \text{LSTM}\left(\mathbf{x}_t, \mathbf{h}_{t-1}, \mathbf{c}_{t-1}\right) \quad (t \geq 1), \\
y_t &= \text{softmax}\left(\mathbf{W}_o \mathbf{h}_t + \mathbf{b}_o\right),
\end{aligned}$$

where CNN(·) is a function that outputs the image features extracted by CNN, that is, the final layer of CNN, and $y_t$ is the $t$th output word. The input $\mathbf{x}_t$ at time $t$ is substituted by a word embedding vector corresponding to the previous output, that is, $y_{t-1}$. The generation process is repeated until LSTM outputs the symbol that indicates the end of sentence.

---

[3] Although their original paper used RNN, they reported in the appendix that LSTM performed better than RNN. Thus, we used LSTM.

In the training phase, given the training data, we train $\mathbf{W}^{(im)}$, $\mathbf{b}^{(im)}$, $\mathbf{W}_*$, $\mathbf{b}_*$, CNN, and LSTM parameters, where $*$ represents wild card.

## 5 Experiments

In this section, we perform an experiment which generates Japanese captions using STAIR Captions. The aim of this experiment is to show the necessity of a Japanese caption dataset. In particular, we evaluate quantitatively and qualitatively how fluent Japanese captions can be generated by using a neural network-based caption generation model trained on STAIR Captions.

### 5.1 Experimental Setup

#### 5.1.1 Evaluation Measure

Following the literature (Chen et al., 2015; Karpathy and Fei-Fei, 2015), we use BLEU (Papineni et al., 2002), ROUGE (Lin, 2004), and CIDEr (Vedantam et al., 2015) as evaluation measures. Although BLEU and ROUGE were developed originally for evaluating machine translation and text summarization, we use them here because they are often used for measuring the quality of caption generation.

#### 5.1.2 Comparison Methods

In this experiment, we evaluate the following caption generation methods.

- En-generator → MT: A pipeline method of English caption generation and English-Japanese machine translation. This method trains a neural network, which generates English captions, with MS-COCO. In the test phase, given an image, we first generate an English caption to the image by the trained neural network, and then translate the generated caption into Japanese one by machine translation. Here, we use Google translate[4] for machine translation. This method is the baseline.

- Ja-generator: This method trains a neural network using STAIR Captions. Unlike MS-COCO → MT, this method directly generate a Japanese caption from a given image .

As mentioned in Section 4, we used the method proposed by Karpathy and Fei-Fei (2015) as caption generation models for both En-generator → MT and Ja-generator.

---

[4] https://translate.google.com/

Table 2: Experimental results of Japanese caption generation. The numbers in boldface indicate the best score for each evaluation measure.

|  | BLEU-1 | BLEU-2 | BLEU-3 | BLEU-4 | ROUGE_L | CIDEr |
| --- | --- | --- | --- | --- | --- | --- |
| En-generator → MT | 0.565 | 0.330 | 0.204 | 0.127 | 0.449 | 0.324 |
| Ja-generator | **0.763** | **0.614** | **0.492** | **0.385** | **0.553** | **0.833** |

Table 3: Examples of generated image captions. En-generator denotes the caption generator trained with MS-COCO. En-generator → MT is the pipeline method: it first generates English caption and performs machine translation subsequently. Ja-generator was trained with Japanese captions.

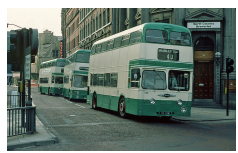

**En-generator:**
A double decker bus driving down a street.
**En-generator → MT:**
ストリートを運転する二重デッカーバス。
**Ja-generator:**
二階建てのバスが道路を走っている。

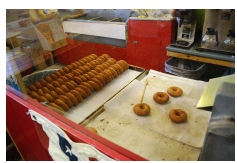

**En-generator:**
A bunch of food that are on a table.
**En-generator → MT:**
テーブルの上にある食べ物の束。
**Ja-generator:**
ドーナツがたくさん並んでいる。

In both the methods, following Karpathy and Fei-Fei, we only trained LSTM parameters, while CNN parameters were fixed. We used VGG with 16 layers as CNN, where the VGG parameters were the pre-trained ones[5]. With the optimization of LSTM, we used mini-batch RMSProp, and the batch size was set to 20.

### 5.1.3 Dataset Separation

Following the experimental setting in the previous studies (Chen et al., 2015; Karpathy and Fei-Fei, 2015), we used 123,287 images included in the MS-COCO training and validation sets and their corresponding Japanese captions. We divided the dataset into three parts, i.e., 113,287 images for the training set, 5,000 images for the validation set, and 5,000 images for the test set.

The hyper-parameters of the neural network were tuned based on CIDEr scores by using the validation set. As preprocessing, we applied morphological analysis to the Japanese captions using MeCab[6].

---
[5] http://www.robots.ox.ac.uk/~vgg/research/very_deep/
[6] http://taku910.github.io/mecab/

### 5.2 Results

Table 2 summarizes the experimental results. The results show that Ja-generator, that is, the approach in which Japanese captions were used as training data, outperformed En-generator → MT, which was trained without Japanese captions.

Table 3 shows two examples where Ja-generator generated appropriate captions, whereas En-generator → MT generated unnatural ones. In the example at the top in Table 3, En-generator first generated the term, "*A double decker bus*." MT translated the term into as "二重デッカーバス", but the translation is word-by-word and inappropriate as a Japanese term. By contrast, Ja-generator generated "二階建てのバス (*two-story bus*)," which is appropriate as the Japanese translation of *A double decker bus*. In the example at the bottom of the table, En-generator → MT yielded the incorrect caption by translating "*A bunch of food*" as "食べ物の束 (*A bundle of food*)." By contrast, Ja-generator correctly recognized that the food pictured in the image is a donut, and expressed it as "ドーナツがたくさん (*A bunch of donuts*)."

## 6 Conclusion

In this paper, we constructed a new Japanese image caption dataset called STAIR Captions. In STAIR Captions, Japanese captions are provided for all the images of MS-COCO. The total number of Japanese captions is 820,310. To the best of our knowledge, STAIR Captions is currently the largest Japanese image caption dataset.

In our experiment, we compared the performance of Japanese caption generation by a neural network-based model with and without STAIR Captions to highlight the necessity of Japanese captions. As a result, we showed the necessity of STAIR Captions. In addition, we confirmed that Japanese captions can be generated simply by adapting the existing caption generation method.

In future work, we will analyze the experimental results in greater detail. Moreover, by using both Japanese and English captions, we will develop multi-lingual caption generation models.